\def\year{2022}\relax
\documentclass[letterpaper]{article} 
\usepackage{aaai22}  
\usepackage{times}  
\usepackage{helvet}  
\usepackage{courier}  
\usepackage[hyphens]{url}  
\usepackage{graphicx} 
\usepackage{natbib}  
\usepackage{caption} 
\DeclareCaptionStyle{ruled}{labelfont=normalfont,labelsep=colon,strut=off} 
\usepackage{algorithm}
\usepackage{algorithmic}

\usepackage{newfloat}
\usepackage{listings}
\floatstyle{ruled}
\newfloat{listing}{tb}{lst}{}
\floatname{listing}{Listing}

\usepackage{subcaption}
\usepackage{booktabs}
\usepackage{multirow}
\usepackage{enumitem}
\usepackage{xcolor}
\usepackage{amsmath}
\usepackage{amssymb}

\newcommand{\xv}{{\boldsymbol x}}
\newcommand{\yv}{{\boldsymbol y}}
\newcommand{\zv}{{\boldsymbol z}}

\newcommand{\mv}{{\boldsymbol m}}
\newcommand{\thetav}{{\boldsymbol \theta}}
\newcommand{\phiv}{{\boldsymbol \phi}}
\newcommand{\deltav}{{\boldsymbol \delta}}
\newcommand{\R}{{\mathbb{R}}}
\newcommand{\E}{{\mathbb{E}}}
\newcommand{\Dcal}{{\mathcal{D}}}
\newcommand{\Tcal}{{\mathcal{T}}}
\newcommand{\Scal}{{\mathcal{S}}}
\newcommand{\Pcal}{{\mathcal{P}}}
\newcommand{\Lcal}{{\mathcal{L}}}
\newcommand{\Rcal}{{\mathcal{R}}}

\usepackage{xcolor}
\definecolor{bittersweet}{rgb}{1.0, 0.44, 0.37}
\definecolor{mygreen}{rgb}{0.29, 0.7, 0.48}

\title{Playing Lottery Tickets with Vision and Language}
\author{
    Zhe Gan,\textsuperscript{\rm 1} Yen-Chun Chen,\textsuperscript{\rm 1} Linjie Li,\textsuperscript{\rm 1} Tianlong Chen,\textsuperscript{\rm 2} \\ 
    Yu Cheng,\textsuperscript{\rm 1} Shuohang Wang,\textsuperscript{\rm 1} Jingjing Liu,\textsuperscript{\rm 3} Lijuan Wang,\textsuperscript{\rm 1} Zicheng Liu\textsuperscript{\rm 1}
}
\affiliations{
    \textsuperscript{\rm 1}Microsoft Corporation, \textsuperscript{\rm 2}University of Texas at Austin, \textsuperscript{\rm 3}Tsinghua University\\
    \{zhe.gan, yen-chun.chen, lindsey.li, yu.cheng, shuowa, lijuanw, zliu\}@microsoft.com \\
    tianlong.chen@utexas.edu, JJLiu@air.tsinghua.edu.cn
    
}

\begin{document}

\maketitle

\begin{abstract}
Large-scale pre-training has recently revolutionized vision-and-language (VL) research. Models such as LXMERT and UNITER have significantly lifted the state of the art over a wide range of VL tasks. However, the large number of parameters in such models hinders their application in practice. In parallel, work on the lottery ticket hypothesis (LTH) has shown that deep neural networks contain small matching subnetworks that can achieve on par or even better performance than the dense networks when trained in isolation. In this work, we perform the first empirical study to assess whether such trainable subnetworks also exist in pre-trained VL models. We use UNITER as the main testbed (also test on LXMERT and ViLT), and consolidate 7 representative VL tasks for experiments, including visual question answering, visual commonsense reasoning, visual entailment, referring expression comprehension, image-text retrieval, GQA, and NLVR$^2$. Through comprehensive analysis, we summarize our main findings as follows. ($i$) It is difficult to find subnetworks that strictly match the performance of the full model. However, we can find ``relaxed'' winning tickets at 50\%-70\% sparsity that maintain 99\% of the full accuracy. ($ii$) Subnetworks found by task-specific pruning transfer reasonably well to the other tasks, while those found on the pre-training tasks at 60\%/70\% sparsity transfer universally, matching 98\%/96\% of the full accuracy on average over all the tasks. ($iii$) Besides UNITER, other models such as LXMERT and ViLT can also play lottery tickets. However, the highest sparsity we can achieve for ViLT is far lower than LXMERT and UNITER (30\% vs. 70\%). ($iv$) LTH also remains relevant when using other training methods (\emph{e.g.}, adversarial training).
\end{abstract}

\section{Introduction}
Inspired by the success of BERT~\cite{devlin2018bert}, vision-and-language pre-training (VLP) has becoming an increasingly central paradigm for vision-and-language (VL) research. Models such as LXMERT~\cite{tan2019lxmert}, ViLBERT~\cite{lu2019vilbert} and UNITER~\cite{chen2020uniter}, have achieved state-of-the-art performance across a wide range of VL tasks, such as visual question answering (VQA)~\cite{antol2015vqa,goyal2017making}, visual commonsense reasoning (VCR)~\cite{zellers2019recognition}, and image-text retrieval~\cite{lee2018stacked}. Despite its empirical success, the memory and computation footprint of these pre-trained models is huge because of their large number of parameters, making it infeasible to use them in resource-constrained scenarios. A natural question that came to our mind: \emph{Can we prune a large pre-trained VL model while preserving its performance and transferability?}

\begin{figure}[!t]
  \centering
  \includegraphics[width=\linewidth]{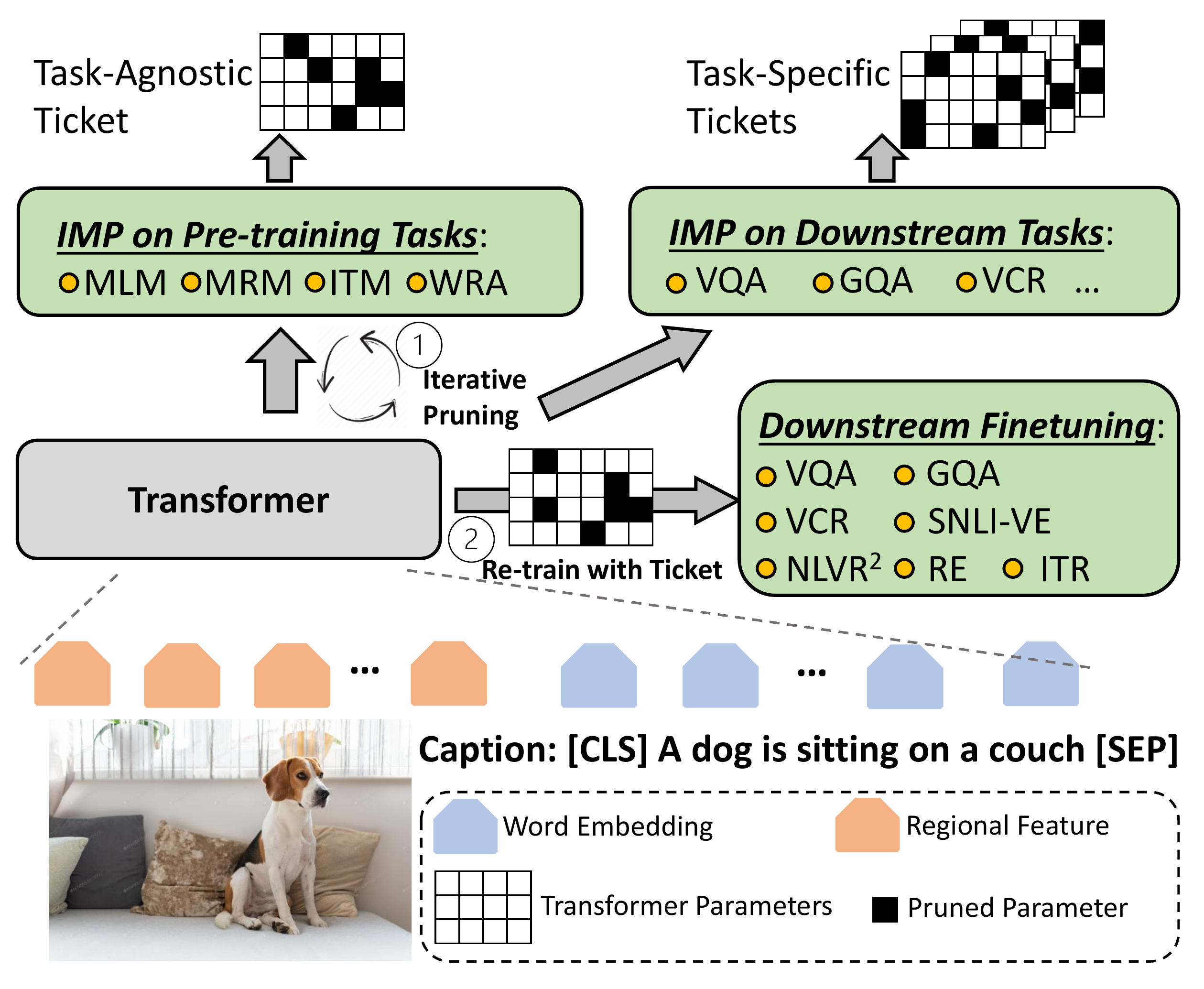}
  \caption{Overview of our training paradigm for \emph{playing lottery tickets with vision and language}. Matching subnetworks (or winning tickets) can be found by Iterative Magnitude-based Pruning (IMP).
  We then re-train the found ticket with the original parameter initialization to verify the downstream performance. Not only \emph{task-specific} winning tickets can be found when running IMP on each downstream task separately, a \emph{task-agnostic} winning ticket is also discovered via IMP on joint pre-training. The task-agnostic ticket results in \emph{universally transferable} subnetworks at 60\%/70\% sparsity that matches 98\%/96\% of the full accuracy averaged over all the tasks considered.}
  \label{fig:framework}
\end{figure}

In this work, we aim to answer this question via the lens of \emph{lottery ticket hypothesis} (LTH)~\cite{frankle2019lottery}, which states that there exist matching subnetworks in dense neural networks that can be trained in isolation from initialization to reach a comparable accuracy to the full model within similar training iterations. LTH has been shown great success in various fields~\cite{yu2019playing,renda2020comparing,chen2020lottery}, and its properties have been widely studied~\cite{malach2020proving,pensia2020optimal,frankle2020linear}. However, LTH has not been introduced to VL tasks yet, it could be a powerful tool to understand the parameter redundancy in the current prevailing VLP models. To start, we use UNITER~\cite{chen2020uniter} as the main testbed, and consider 7 representative VL tasks for experiments, including VQA~\cite{goyal2017making}, VCR~\cite{zellers2019recognition}, GQA~\cite{hudson2019gqa}, NLVR$^2$~\cite{suhr2018corpus}, visual entailment~\cite{xie2019visual}, referring expression comprehension~\cite{yu2016modeling}, and image-text retrieval~\cite{lee2018stacked}. In our context, a \emph{ticket} means a VLP subnetwork, and a \emph{winning ticket} means a subnetwork that can match the performance of the original full VLP model. Based upon this, we ask the following three questions: 
\begin{itemize}[leftmargin=*]
    \item \textbf{\emph{Existence}}: Can we draw winning tickets successfully for various VL tasks?
    \item \textbf{\emph{Transferability}}: Can we find tickets that transfer universally to all downstream VL tasks?
    \item \textbf{\emph{Compatibility}}: Do the LTH observations still hold when switching to different backbones (\emph{e.g.}, LXMERT~\cite{tan2019lxmert}, ViLT~\cite{kim2021vilt}), and training strategies (\emph{e.g.}, adversarial training)?
\end{itemize}
First, \emph{can we draw VL winning tickets?} To answer this, we use the pre-trained weights as our model initialization for task-specific finetuning, and use Iterative Magnitude-based Pruning (IMP)~\cite{han2015deep} to draw the tickets for each VL task. However, finding tickets through iterative and repeated train-prune-retrain cycles for each task is very time-consuming, primarily when a large pre-trained model is used here. Then, it becomes critical to ask: \emph{how can we find subnetworks that transfer universally?} If this can be achieved, the extraordinary cost of finding a winning ticket can be amortized by transferring it to a range of downstream tasks. Inspired by~\citet{chen2020lottery}, a natural idea is to perform IMP on the pre-training tasks using the pre-training data, and assess whether such learned tickets are transferable or not,  since pre-training can be considered as task-agnostic.
Besides this, we further comprehensively analyze the transfer behavior among all the downstream tasks to  better  understand the found task-specific winning tickets.

The above analysis is conducted on UNITER, which is a one-stream model and uses an object detection module to first extract visual features offline. To study the compatibility of LTH, we also experiment on LXMERT (a two-stream model instead), and ViLT (directly taking image patches and word tokens as model inputs). Moreover, instead of cross-entropy training, we further test LTH under adversarial training~\cite{gan2020large} to investigate its corresponding training behaviors. Through comprehensive analysis, we summarize our main findings as follows. 
\begin{itemize}[leftmargin=*]
    \item \textbf{\emph{VLP can play lottery tickets too}}: It is difficult to find UNITER subnetworks that \emph{strictly} match the full performance, even with rewinding. However, it is encouraging that \emph{``relaxed''} winning tickets that match 99\% of the full accuracy can be found at 50\%-70\% sparsity across all the VL tasks considered.  
    \item \textbf{\emph{One ticket to win them all}}: Matching subnetworks found via IMP on pre-training tasks transfer universally. Interestingly, matching subnetworks found on each downstream task also transfer to other tasks well, indicating that the learned task-specific subnetworks do not aggressively overfit to one specific task.
    \item \textbf{\emph{Different VLP models behave differently}}: Though all the VLP models can play lottery tickets, we also observe that the highest sparsity we can achieve for ViLT is far lower than LXMERT and UNITER (30\% vs. 70\%). 
    \item \textbf{\emph{Playing lottery tickets adversarially}}: Compared with standard cross-entropy training, we observe that sparse winning tickets can also be identified with adversarial training, with enhanced performance.
\end{itemize}
We conclude that the primary LTH observations found in computer vision, NLP, and other areas also hold in the context of vision and language.  

\section{Related Work}
\paragraph{Vision-and-Language Pre-training (VLP).}
The past two years have witnessed a boom of VLP methods.  
By adopting transformer~\cite{vaswani2017attention} as the building block, early approaches use a two-stream architecture for multimodal fusion~\cite{lu2019vilbert,tan2019lxmert,lu201912}, while single-stream architecture has gained popularity later on~\cite{su2019vl,li2019visualbert,li2019unicoder,chen2020uniter,zhou2019unified,gan2020large,li2020oscar,zhang2021vinvl}. 
While most of these methods rely on an object detection module to extract visual features offline, recently, end-to-end VLP methods~\cite{huang2020pixel,huang2021seeing,kim2021vilt,xue2021probing,li2021align,dou2021empirical} are becoming increasingly popular.

Different from these efforts on making VLP models larger and stronger, we focus on a different direction, making VLP models \emph{smaller}. Note that two recent works, MiniVLM~\cite{wang2020minivlm} and DistilVLM~\cite{fang2021compressing}, have also attempted to train a smaller VLP model; however, our focus is different from theirs. Specifically, MiniVLM directly adopts MiniLM~\cite{wang2020minilm} for the transformer
module, while spending a larger portion of efforts on designing a compact image feature extractor; DistilVLM focuses on knowledge distillation. Here, we study the over-parameterization of VLP models via the lens of \emph{lottery ticket hypothesis}, a popular concept in deep learning nowadays, but not introduced to VL research yet. 

\paragraph{Lottery Ticket Hypothesis (LTH).}
LTH~\cite{frankle2019lottery} claims the existence of sparse, separate trainable subnetworks that are able to match or even surpass the performance of the original dense network. Though originally working only on small networks, later on, rewinding is found to be a useful technique to scale up LTH to large networks~\cite{renda2020comparing,frankle2020linear}. Since its birth, LTH has received wide attention and becomes an emerging subfield in deep learning. The properties of LTH are widely studied for image classification~\cite{liu2018rethinking,evci2019difficulty,Frankle2020The,savarese2020winning,grasp,You2020Drawing,ma2021good}. Recently, LTH has also been evidenced across other fields, such as NLP~\cite{gale2019state,yu2019playing,prasanna2020bert,chen2020lottery,chen2020earlybert}, object detection~\cite{girish2020lottery}, generative adversarial networks~\cite{chen2021gans,kalibhat2020winning,chen2021ultra}, graph neural networks~\cite{chen2021unified}, reinforcement learning~\cite{yu2019playing}, and life-long learning~\cite{chen2021long}. 

Recent work has also started to investigate the existence of winning tickets in self-supervised pre-training of visual encoders~\cite{chen2020lottery2} and language models~\cite{chen2020lottery,chen2020earlybert}.
However, to the best of our knowledge, the study of lottery tickets in VLP remains untouched. As VLP becomes increasingly popular, it is critical to understand the parameter redundancy in such models, potentially making them small without sacrificing the performance. 

\section{Preliminaries}
In this section, we detail the techniques we use to identify winning tickets, and present our setup for empirical study.

\paragraph{Backbones.} 
We use UNITER~\cite{chen2020uniter} as an example to introduce the backbone, which shares the same structure as BERT, except that the input is a mixed sequence of two modalities.
Specifically, given a dataset that consists of image-text pairs $\xv=(\xv_{img}, \xv_{txt})$, UNITER first encodes the corresponding image regions and textual tokens into low-dimensional feature vectors $\zv_{img}=g_{bu}(\xv_{img})$ and $\zv_{txt}=g_{emb}(\xv_{txt})$, where $g_{bu}(\cdot)$ is the fixed bottom-up image feature extractor~\cite{anderson2018bottom}, $g_{emb}(\cdot)$ is a learnable word embedding function. Then, a transformer is applied on top to obtain contextualized representations: $\tilde{\zv}_{img}, \tilde{\zv}_{txt}, \tilde{\zv}_{cls} = f_1({\xv_{img}, \xv_{txt}}; \thetav)$, where a special \texttt{[CLS]} token is employed whose embedding $\tilde{\zv}_{cls}$ is considered as the joint multimodal representation. $\thetav \in \R^{d_1}$ includes all the trainable parameters.  For a particular downstream task, we add a final, task-specific classification layer on top of $\tilde{\zv}_{cls}$ to obtain the output logit vector $f_2(\tilde{\zv}_{cls}; \phiv)$, where $\phiv\in \R^{d_2}$ denotes task-specific parameters. The whole UNITER network is abbreviated as $f(\xv; \thetav, \phiv)$ that absorbs both $f_1(\cdot,\cdot)$ and $f_2(\cdot)$. For LXMERT~\cite{tan2019lxmert}, it takes the same image features from object detection as model input, but adopts a two-stream model architecture instead. For ViLT~\cite{kim2021vilt}, it uses the same one-stream architecture, but directly takes image patches and word tokens as inputs, and models all the intra- and inter-modality interaction via a single unified transformer. 

Given the task-specific supervision signal $\yv$ (typically a label in VL tasks), model training can be summarized as:
\begin{align} \label{eqn:std_xe}
    \min_{\thetav,\phiv} \E_{(\xv, \yv)\sim \Dcal} [L(f(\xv;\thetav,\phiv),\yv)]\,,
\end{align}
where $L(\cdot)$ is the cross-entropy loss, and $\Dcal$ denotes the dataset for a downstream task. 
We use the official UNITER/LXMERT/ViLT code bases for experiments. 

\paragraph{Subnetworks.}
A subnetwork of $f(\xv; \thetav, \phiv)$ means a network $f(\xv; \mv \odot \thetav, \phiv)$ with a binary pruning mask $\mv \in \{0,1\}^{d_1}$ indicating which part of the parameters are set to 0, and $\odot$ is the element-wise product. Following~\citet{frankle2019lottery}, we define a \emph{matching subnetwork} as a subnetwork that can be trained to the full accuracy of the dense network within similar training iterations. A \emph{winning ticket} is defined as a matching subnetwork $f(\xv;\mv \odot \thetav_0, \cdot)$ where $\thetav=\thetav_0$, which is typically is a random weight initialization. However, in our context, $\thetav_0$ represents the pre-trained model weights. We also define a ``\emph{relaxed}'' winning ticket as one that matches $p\%$ of the the full accuracy, where $p$ is set to a large number close to $100$ (such as $99$). 

\paragraph{Finding Subnetworks.}
As used in many lottery ticket papers, we use Iterative Magnitude-based Pruning (IMP)~\cite{han2015deep} to find the subnetwork. Specifically, the pruning mask $\mv$ is determined by training the unpruned network to completion on a downstream task, then pruning individual weights with the lowest magnitudes globally throughout the network. The weights are then reset to the pre-trained initialization $\thetav_0$ (or, $\thetav_i$ for a specific \emph{rewinding} step $i$ in training), and only the learned mask $\mv$ is stored. We prune a certain amount (\emph{e.g.}, 10\%) of non-zero weights after completion, and re-train the network several times to meet the sparsity requirement. The full IMP procedure is provided in the Appendix.

We consider finding subnetworks via both ($i$) task-specific finetuning and ($ii$) task-agnostic pre-training,\footnote{We only perform pre-training on UNITER, since pre-training is heavy; we perform finetuning for UNITER, LXMERT, and ViLT.} hoping that universal transferable subnetworks can be identified. For UNITER pre-training, we use all the pre-training tasks to learn the mask, including Masked Language Modeling, Masked Region Modeling, Image-Text Matching, and Word-Region Alignment. See~\citet{chen2020uniter} for details of these tasks. As our model is initialized by pre-trained UNITER, we further pre-train only 10\% of original training steps in each pruning round (we prune 9 rounds in total). Therefore, the total time spent for a full IMP process roughly equals the time used for pre-training UNITER from scratch.

\paragraph{Evaluation of Subnetworks.}
For a particular downstream task, after obtaining a subnetwork $f(\xv; \mv \odot \thetav, \cdot)$, we reset the weights to $\thetav_0$ or $\thetav_i$ (if rewinding is used), and then completely re-train the subnetwork to test whether the final subnetworks can still achieve the original accuracy. 
For pre-training, since the performance of the pre-training tasks validation loss does not correlate to the task-specific performance~\cite{chen2020uniter}, we finetune and test the identified subnetworks on all the downstream tasks. We use both the in-domain and out-of-domain image-text datasets for IMP-based pre-training, including COCO~\cite{lin2014microsoft}, Visual Genome~\cite{krishna2017visual}, Conceptual Captions~\cite{sharma2018conceptual}, and SBU Captions~\cite{ordonez2011im2text}.   

\begin{table*}[!t]
\small
\resizebox{1.0\textwidth}{!}
{
\begin{tabular}{cc|cccccccc}
& \multirow{2}{*}{Dataset} & VQA & GQA & VCR & NLVR$^2$ & SNLI-VE & RefCOCO+ & Flickr30k IR & Flickr30k TR  \\ 
& & mini-dev$^\dagger$ & test-dev & Q$\rightarrow$AR val & dev & val & val$^d$ & R@1 & R@1 \\ \cline{3-10}
\# &Sparsity & 70\% & 70\% & 50\% & 60\% & 60\% & 70\% & 60\% & 60\% \\
\hline
1 &UNITER$_\text{B}$ (paper) & 70.75 & $-$ & 54.94 &77.18 &78.59 & 75.31 & 72.52 & 85.90 \\
2 & UNITER$_\text{B}$ (reimp.) & 70.64$\pm$\tiny{0.06} & 59.64$\pm$\tiny{0.15} & 54.37$\pm$\tiny{0.31}\small{$^\ddagger$} & 76.75$\pm$\tiny{0.19} & 78.47$\pm$\tiny{0.10} & 74.73$\pm$\tiny{0.06} & 71.25$\pm$\tiny{0.11}\small{$^\star$} & 84.63$\pm$\tiny{1.02}\small{$^\star$} \\
3 & $\times 99\%$ & 69.93 & 59.04 & 53.83 & 75.98 & 77.69 & 73.98 & 70.54 & 83.78\\ 
\hline
4 & $f(\xv;\mv_{\text{IMP}} \cdot \thetav_0)$ & 69.98$\pm$\tiny{0.05} & 59.26$\pm$\tiny{0.09} & 53.15$\pm$\tiny{1.02} & 76.32$\pm$\tiny{0.41} & 77.69$\pm$\tiny{0.07} & 74.06$\pm$\tiny{0.27} & 70.15$\pm$\tiny{0.71} & 83.77$\pm$\tiny{0.76} \\
5 & $f(\xv;\mv_{\text{RP}} \cdot \thetav_0)$ & 60.45 & 55.95 & 25.35 & 52.42 & 71.30 & 72.95 & 61.44 & 76.80 \\
6 & $f(\xv;\mv_{\text{IMP}} \cdot \thetav_0')$ & 67.98 & 58.45 & 50.39 & 54.15 & 76.45 & 71.09 & 63.38 & 79.30 \\
7 & $f(\xv;\mv_{\text{IMP}} \cdot \thetav_0'')$ & 60.46 & 47.49 & 6.25 & 51.52 & 69.32 & 67.34 & 38.94 & 48.00 \\
\end{tabular}
}
\caption{
Performance of subnetworks at the highest sparsity for which IMP finds ``relaxed'' winning tickets that maintains 99\% of the full accuracy on each task. Entries with $\pm$ are the average across three runs. IMP: Iterative Magnitude Pruning; RP: Random Pruning; $\thetav_0$: pre-trained UNITER weights; $\thetav_0'$: pre-trained BERT weights; $\thetav_0''$: randomly shuffled pre-trained UNITER weights. ($\dagger$) To avoid submitting results to the VQA test server too frequently, instead of reporting results on test-dev/-std sets, we use a mini-dev set for comparison. The same min-dev set was also used in UNITER. ($\ddagger$) For fair comparison on transfer learning, we did not perform 2-nd stage pre-training for VCR task as in UNITER. ($\star$) To rule out other factors that may influence results besides pruning, we did not use hard negative mining as in UNITER.   
}
\label{tab:uniter_lottery_results}
\end{table*}

\paragraph{Downstream Tasks.}
We consider 7 VL tasks for experiments. ($i$) For VQA~\cite{goyal2017making}, GQA~\cite{hudson2019gqa} and VCR~\cite{zellers2019recognition}, given an image and an input question, the model selects an answer from a candidate pool. ($ii$) For NLVR$^2$~\cite{suhr2018corpus}, given a pair of images and a natural language description, the model judges the correctness of the description based on the input image pair. For Visual Entailment~\cite{xie2019visual}, the model predicts whether a given image entails a given sentence. ($iii$) For Referring Expression (RE) Comprehension, we evaluate on RefCOCO+~\cite{yu2016modeling}, where given a text description, the model selects the described region
from a set of image region proposals. ($iv$) For Image-Text Retrieval (ITR), we consider both image retrieval and text retrieval on Flickr30k dataset.

For VCR, 2nd-stage pre-training was found useful in UNITER finetuning. For simplicity and ease of study of transfer learning, we do not use 2nd-stage pre-training. For ITR, hard negative mining is necessary to boost performance. We do not use this as it is computationally heavy, and we aim to study LTH rather than chasing state-of-the-art performance. For VQA, we mainly report results on an internal mini-dev set for faster evaluation of the found  tickets, and avoid submitting results to the VQA test server too frequently. This same mini-dev set is also used in UNITER~\cite{chen2020uniter}. 

\section{Experiments}
In this section, we perform extensive experiments to examine the LTH in the context of vision and language. 

\subsection{VLP Can Play Lottery Tickets Too} \label{sec:task-specific pruning}
First, we evaluate whether winning tickets exist in UNITER. In particular, we answer the following questions.

\paragraph{\emph{Q1: Are there winning tickets in UNITER?}}
To answer this, we first run IMP on a downstream task $\Tcal$ to obtain a sparsity pattern $\mv_{\text{IMP}}^\Tcal$. This produces a subnetwork $f(\xv;\mv_{\text{IMP}}^\Tcal \odot \thetav_0,\cdot)$. We then train this subnetwork again on task $\Tcal$ to evaluate whether this is a winning ticket. 

\begin{figure*}[t]
     \centering
     \begin{subfigure}[b]{0.32\textwidth}
         \centering
         \includegraphics[width=0.98\textwidth]{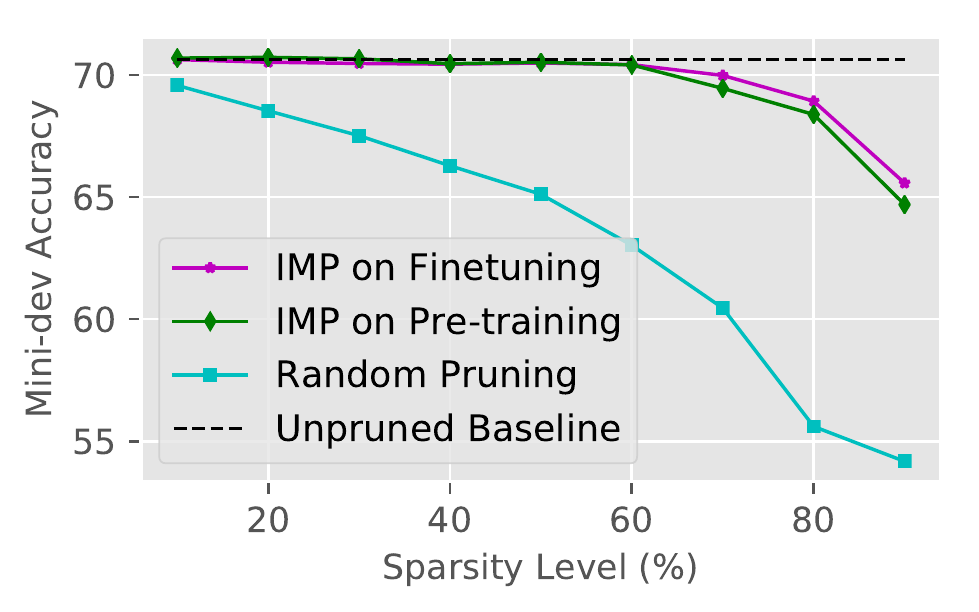}
         \caption{VQA}
     \end{subfigure}
     \begin{subfigure}[b]{0.32\textwidth}
         \centering
         \includegraphics[width=0.98\textwidth]{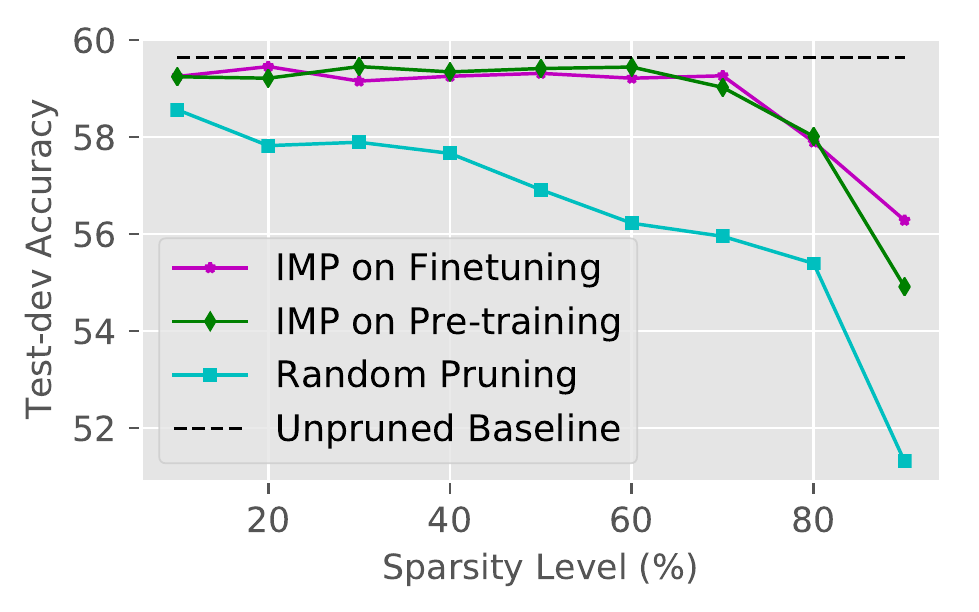}
         \caption{GQA}
     \end{subfigure}
     \begin{subfigure}[b]{0.32\textwidth}
         \centering
         \includegraphics[width=0.98\textwidth]{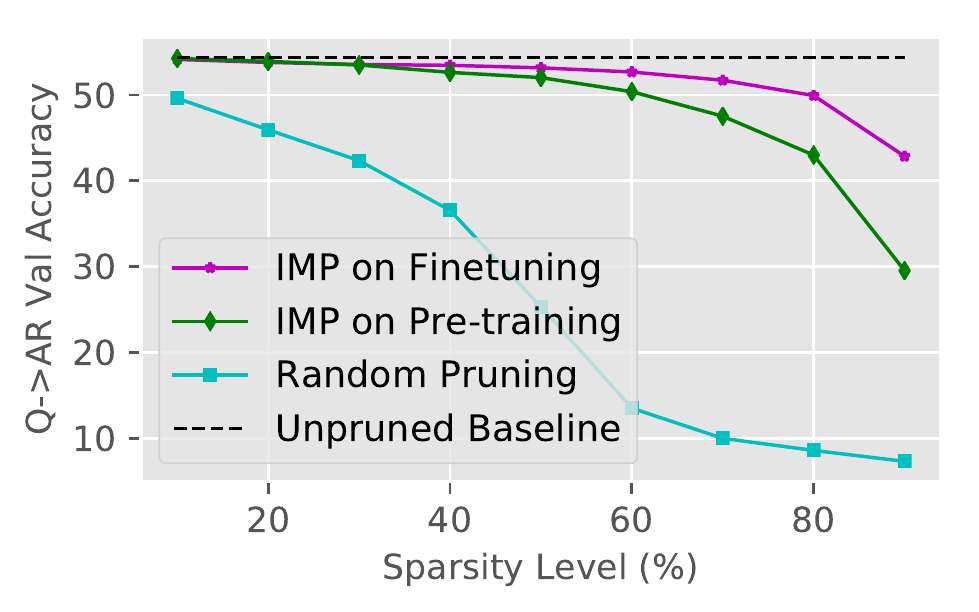}
         \caption{VCR}
     \end{subfigure} \\
     \begin{subfigure}[b]{0.32\textwidth}
         \centering
         \includegraphics[width=0.98\textwidth]{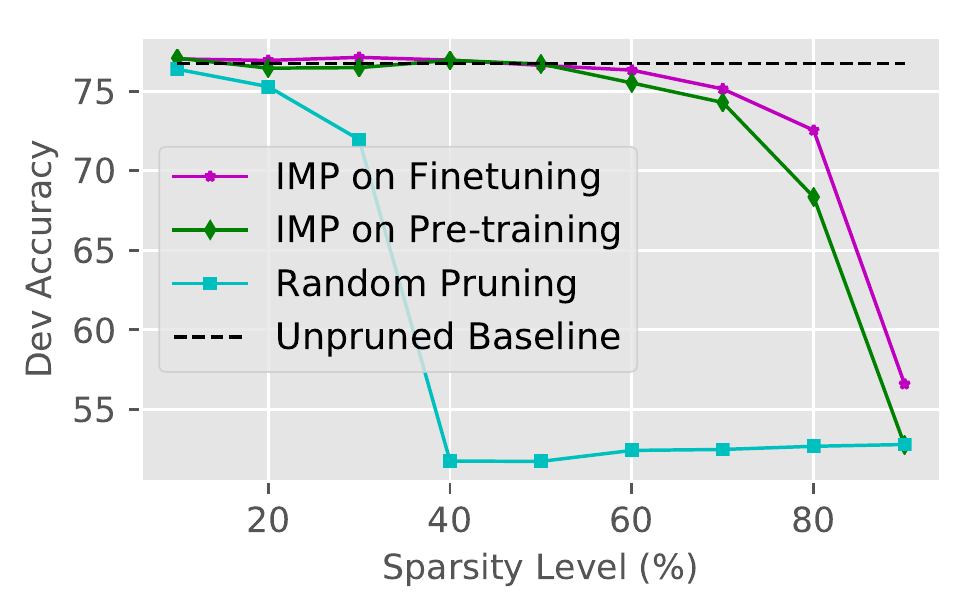}
         \caption{NLVR$^2$}
     \end{subfigure}
     \begin{subfigure}[b]{0.32\textwidth}
         \centering
         \includegraphics[width=0.98\textwidth]{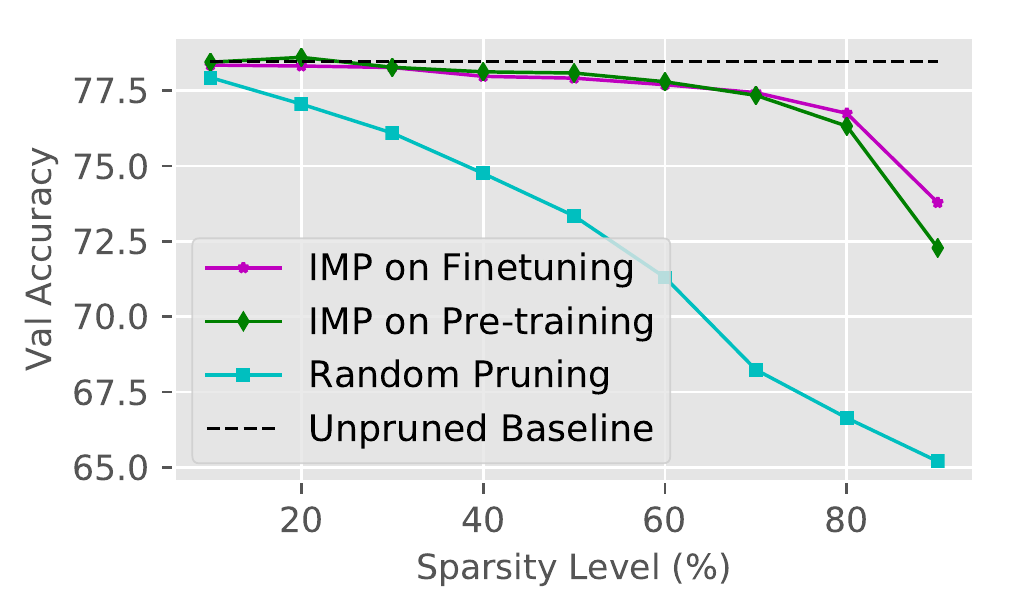}
         \caption{SNLI-VE}
     \end{subfigure}
     \begin{subfigure}[b]{0.32\textwidth}
         \centering
         \includegraphics[width=0.98\textwidth]{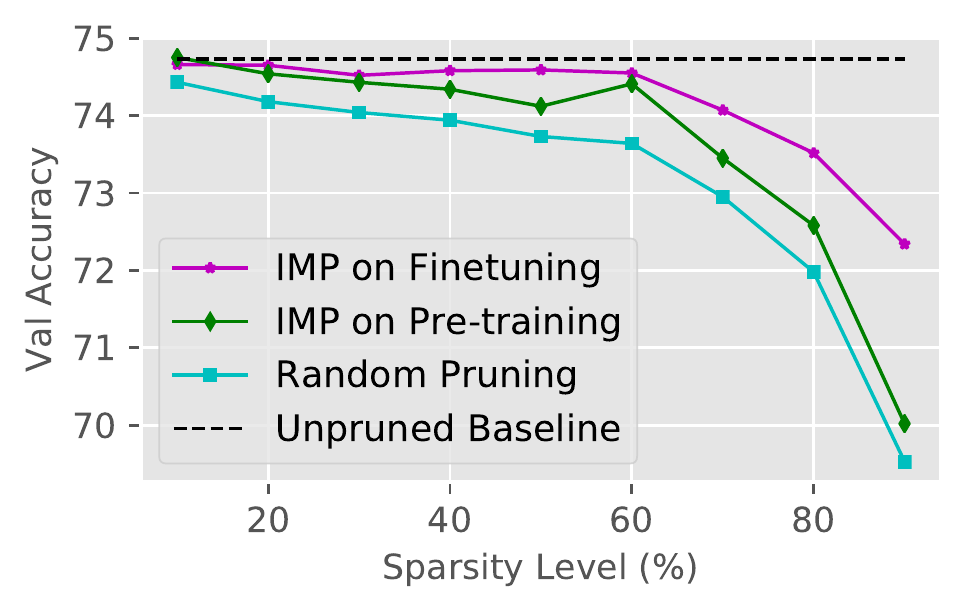}
         \caption{RefCOCO+}
     \end{subfigure} \\
     \begin{subfigure}[b]{0.32\textwidth}
         \centering
         \includegraphics[width=0.98\textwidth]{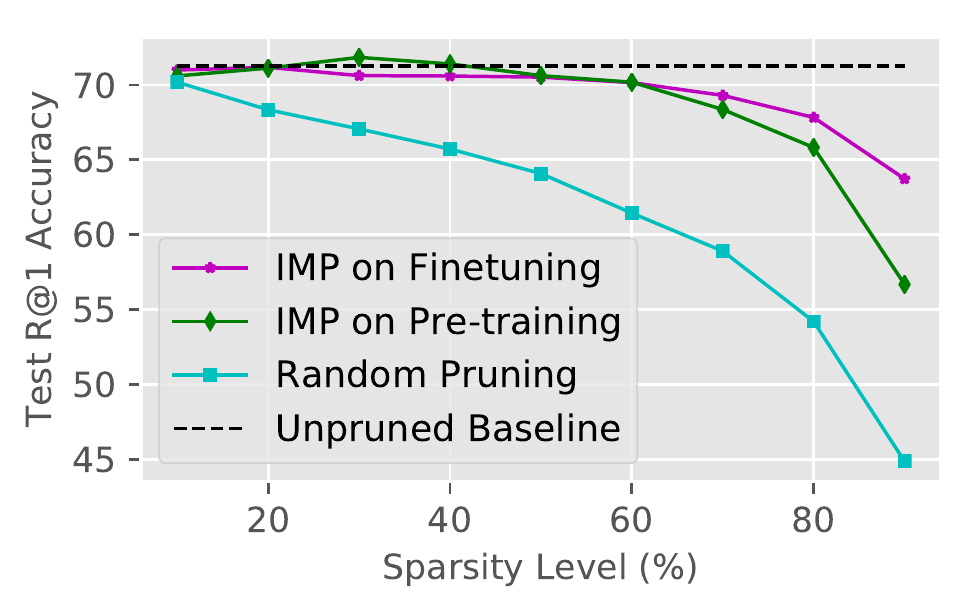}
         \caption{Flickr30k IR}
     \end{subfigure}
     \begin{subfigure}[b]{0.32\textwidth}
         \centering
         \includegraphics[width=0.98\textwidth]{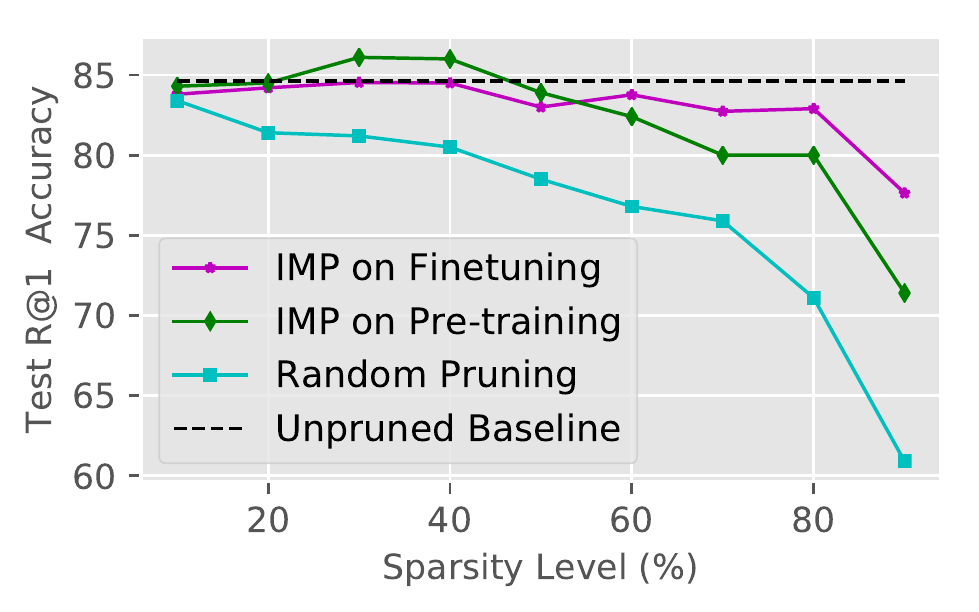}
         \caption{Flickr30k TR}
     \end{subfigure}
     \begin{subfigure}[b]{0.32\textwidth}
         \centering
         \includegraphics[width=0.98\textwidth]{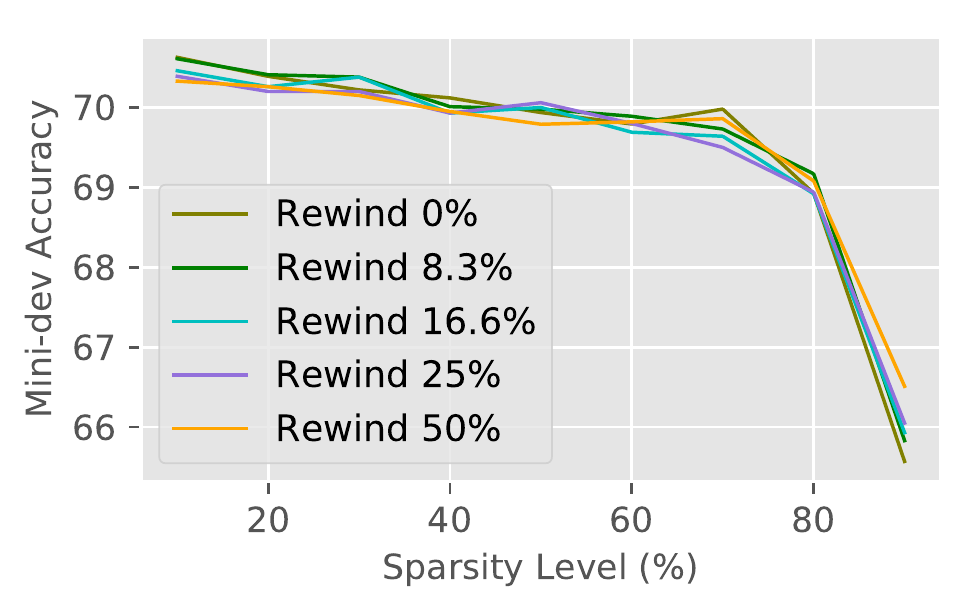}
         \caption{VQA Rewinding}
     \end{subfigure}
     \caption{Comparison among ($i$) IMP performed on task-specific finetuning, ($ii$) IMP performed on task-agnostic pre-training, and ($iii$) random pruning on task-specific finetuning across sparsities for all the tasks. We also report rewinding in the VQA task in sub-figure (i).}
     \label{fig:IMP_vs_RP}
\end{figure*}

Results across all the sparsity levels (10\% to 90\%) on all the downstream tasks are shown in Figure~\ref{fig:IMP_vs_RP} (\textcolor{magenta}{magenta curves}). For tasks of image-text retrieval and NLVR$^2$, matching subnetworks with sparsity 40\% can be identified. However, it is generally challenging to find subnetworks that ``strictly'' match the performance of the full accuracy on the other tasks. Therefore, we define ``\emph{relaxed}'' winning tickets as the ones that can match 99\% of the full accuracy. It will still be encouraging if such subnetworks can be found.

Results are summarized in Table~\ref{tab:uniter_lottery_results}. Row \#1 reports the full UNITER$_{\text{B}}$ performance reported in the UNITER paper~\cite{chen2020uniter}. Row \#2 reports the results of our re-implementation, where different random seeds are used to account for fluctuations. We use the default hyper-parameters provided in the UNITER code base without any tuning. Row \#3 calculates 99\% of the full accuracy on each task for reference. As can be seen from Row \#4, on all VL tasks, ``relaxed'' winning tickets can be found. The highest sparsities range from 50\% (\emph{e.g.}, VCR) to 70\% (\emph{e.g.}, VQA). For VCR, it is challenging to find high-sparsity subnetworks. We hypothesize that commonsense knowledge is harder to learn, and smaller weights also play essential roles in improving model's commonsense reasoning abilities, making the subnetwork for VCR harder to prune.

\paragraph{\emph{Q2: Are winning tickets sparser than randomly pruned or initialized subnetworks?}} Previous work has shown that both the specific learned sparse mask and the specific initialization are necessary for finding winning tickets~\cite{frankle2019lottery}. To assess the importance of the learned mask in the context of UNITER, we compare with a random pruning baseline, and report results in Row \#5 of Table~\ref{tab:uniter_lottery_results}. That is, we finetune a randomly pruned UNITER model on each downstream task. Interestingly, for some tasks (\emph{e.g.}, GQA and RefCOCO+), random pruning achieves pretty strong performance. However, by comparing performance across the board, it is also clear that random pruning performs far worse than the identified winning tickets. In Figure~\ref{fig:IMP_vs_RP}, we further compare IMP and random pruning across all sparsities. Again, random pruning achieves far lower performance, confirming that the sparse structure found by IMP is crucial for the good performance of subnetworks. 

To assess the importance of the initialization, we consider two different initializations with the learned mask unchanged: ($i$) using pre-trained BERT weights $\thetav_0'$ as initialization, and ($ii$) shuffling the UNITER pre-trained weights within each layer to obtain a new initialization $\thetav_0''$. Results of these two baselines are summarized in Row \#6 and \#7 of Table~\ref{tab:uniter_lottery_results}, respectively. Clearly, training from $\thetav_0''$ achieves far lower performance than training from $\thetav_0$. However, it is also interesting to observe that training from $\thetav_0'$ achieves much more reasonable performance, though still lagging behind training from $\thetav_0$, indicating the importance of the specific initialization. We hypothesize the good performance of $\thetav_0'$ is partially due to that $\thetav_0'$ is used as the initialization to pre-train UNITER; therefore, the structure of the UNITER weights may be partially inherited from BERT. 

\noindent\textbf{\emph{Q3: Does rewinding improve performance?}} For large networks, \emph{rewinding} is found to be necessary to identify winning tickets~\cite{renda2020comparing}. After obtaining the masks, instead of resetting the weights to $\thetav_0$, one should rewind the weights to $\thetav_i$, the weights after $i$ steps of training. To examine whether rewinding is helpful in the context of UNITER, we run experiments at different rewinding ratios using VQA as the representative task. Results are shown in Figure~\ref{fig:IMP_vs_RP}(i). Rewinding does not have a notable effect on the VQA performance, with only minor performance improvement observed at high-sparsity ratio (90\%). Similar observations are also found on other downstream tasks. 

\begin{figure*}[t]
     \centering
     \begin{subfigure}[b]{0.32\textwidth}
         \centering
         \includegraphics[width=0.98\textwidth]{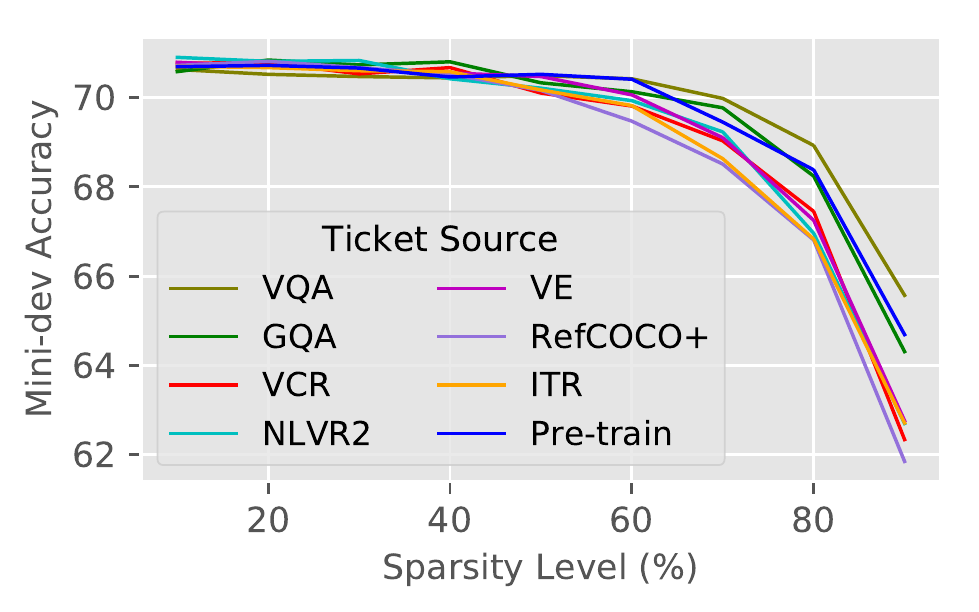}
         \caption{VQA}
     \end{subfigure}
     \begin{subfigure}[b]{0.32\textwidth}
         \centering
         \includegraphics[width=0.98\textwidth]{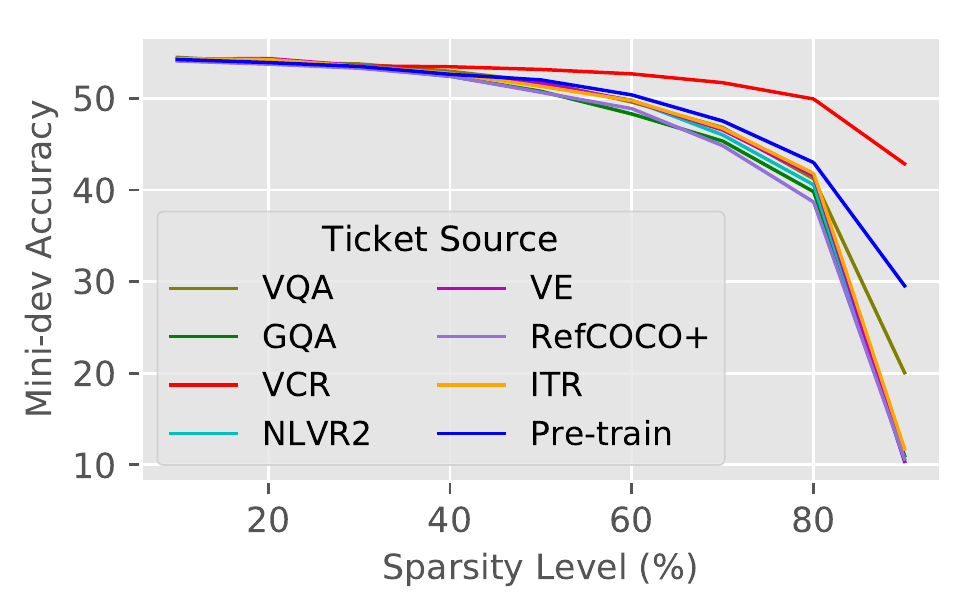}
         \caption{VCR}
     \end{subfigure}
     \begin{subfigure}[b]{0.32\textwidth}
         \centering
         \includegraphics[width=0.98\textwidth]{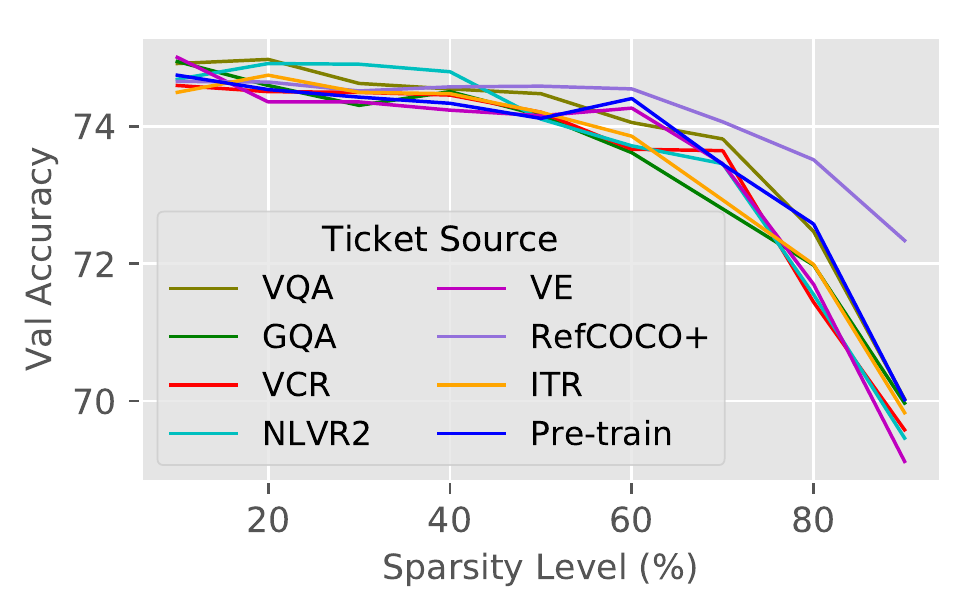}
         \caption{RefCOCO+}
    \end{subfigure}
    \begin{subfigure}[b]{0.32\textwidth}
         \centering
         \includegraphics[width=0.98\textwidth]{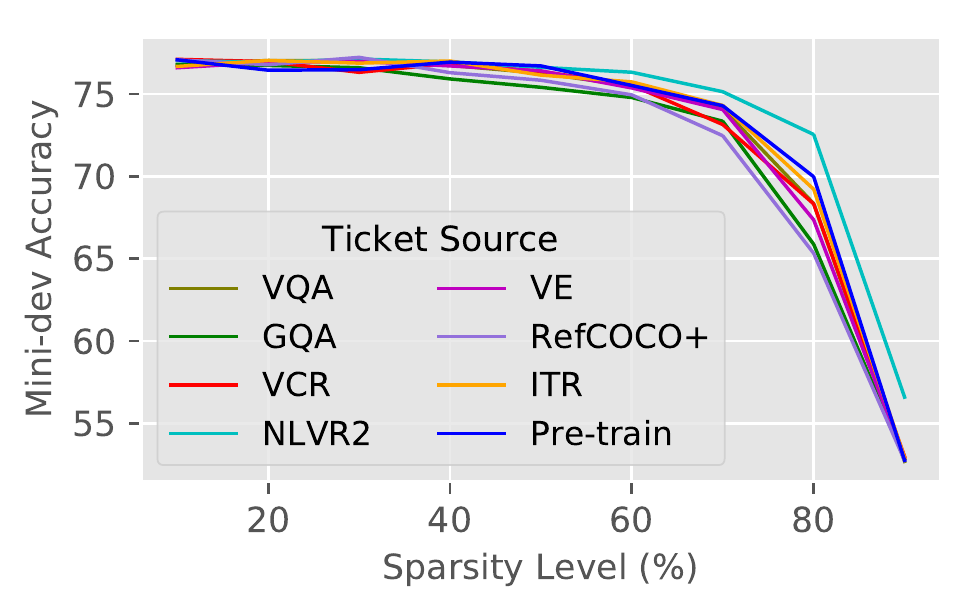}
         \caption{NLVR$^2$}
     \end{subfigure}
     \begin{subfigure}[b]{0.32\textwidth}
         \centering
         \includegraphics[width=0.98\textwidth]{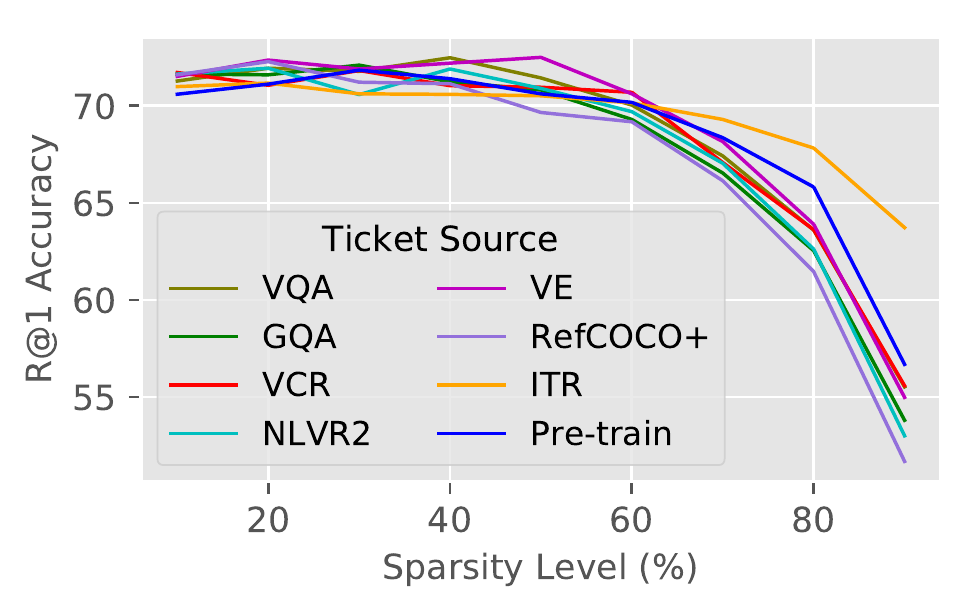}
         \caption{Flickr30k IR}
     \end{subfigure}
     \begin{subfigure}[b]{0.32\textwidth}
         \centering
         \includegraphics[width=0.98\textwidth]{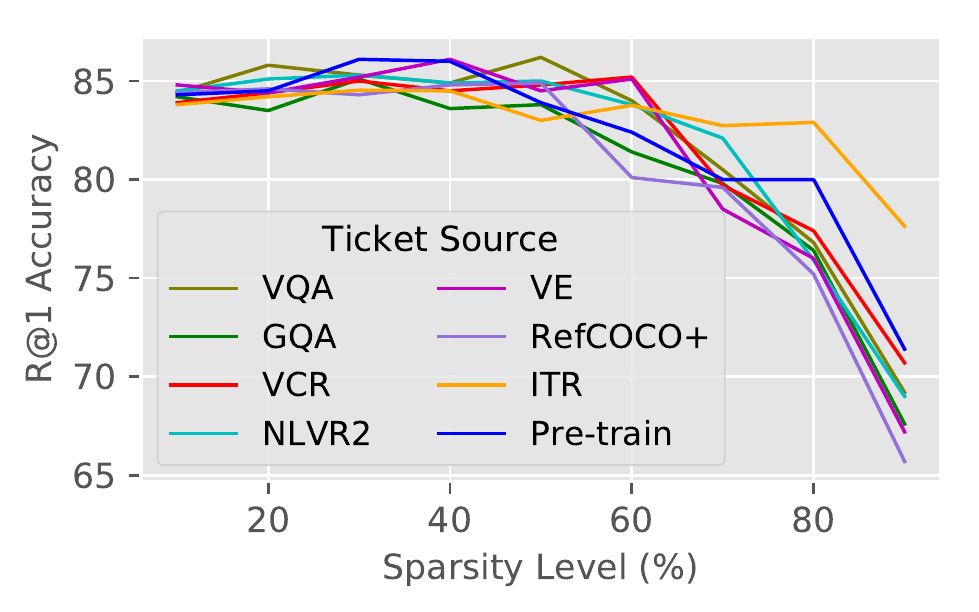}
         \caption{Flickr30k TR}
     \end{subfigure}
     \caption{Transferring winning tickets across tasks. Winning ticket performance on target tasks: (a) VQA, (b) VCR, (c) RefCOCO+, (d) NLVR$^2$, (e) Flickr30k IR, (f) Flickr30k TR. Within each plot, each line represents a different source task for the winning ticket. Better zoomed in and viewed in color. Additional curves are in the Appendix.}
     \label{fig:trans_study}
\end{figure*}

\begin{table*}[!t]
\small
\resizebox{1.0\textwidth}{!}
{
\begin{tabular}{c|cccccccc|cc}
\multirow{2}{*}{Sparisty} & VQA & GQA & VCR & NLVR$^2$ & SNLI-VE & RefCOCO+ & Flickr30k IR & Flickr30k TR &   \multicolumn{2}{c}{  Ave. Perf. Drop (\%)}  \\ \cline{2-11}
& mini-dev & test-dev & Q$\rightarrow$AR val & dev & val & val$^d$ & R@1 & R@1 & All & w/o VCR \\
\hline
0\% & 70.64 & 59.64 & 54.37 & 76.75 & 78.47 & 74.73 & 71.25 & 84.63 & $-$ & $-$ \\
50\%  & 70.52 & 59.41 & 52.01 & 76.71 & 78.08 & 74.12 & 70.62 & 83.90 & 1.00 & 0.52\\
60\%  & 70.41 & 59.44 & 50.37 & 75.52 & 77.79 & 74.41 & 70.18 & 82.40 & 1.88 & 1.10\\
70\%  & 69.45 & 59.02 & 47.52 & 74.29 & 77.34 & 73.45 & 68.36 & 80.00 & 3.90 & 2.66\\
80\%  & 68.38 & 58.01 & 42.99 & 69.98 & 76.32 & 72.58 & 65.82 & 80.00 & 6.80 & 4.78\\
\end{tabular}
}
\caption{
Performance of the universal transferable subnetwork found on pre-training at specified sparsities. 
}
\label{tab:universal_subnetwork_results}
\end{table*}

\subsection{One Ticket to Win Them All} \label{sec:task-agnostic pruning}
Finding winning tickets on each downstream task separately is time-consuming, as each time when IMP is performed, it has to go through the full train-prune-retrain cycle multiple times. In this section, we aim to identify subnetworks that transfer well across all the VL tasks. In particular, we answer the following questions. 

\paragraph{\emph{Q4: Do winning tickets found on pre-training tasks transfer?}} Pre-training is believed to learn \emph{universal} VL representations. As shown in~\citet{cao2020behind}, the pre-trained weights indeed have captured rich visual coreference and visual relation knowledge. This naturally leads to our hypothesis: can the subnetwork identified by the pre-training tasks on the pre-training data also transfer universally? 

To study this, we first identify a subnetwork $f(\xv;\mv_{\text{IMP}}^{\Pcal} \cdot \thetav_0,\cdot)$ on the pre-training tasks $\Tcal$, and then train it on all the downstream tasks to evaluate its performance. Results are summarized in Figure~\ref{fig:IMP_vs_RP} (\textcolor{green}{green curves}). Interestingly, though 
pre-training never obtains the supervision signal in the downstream tasks, the found subnetwork transfers pretty \emph{universally}; only when the sparsity is high (\emph{e.g.}, 80\%, 90\%), the found subnetwork performs worse than the ones found by task-specific IMP, indicating that the pre-training tasks are strong signals for learning how to prune.

\begin{figure*}[t]
     \centering
     \begin{subfigure}[b]{0.32\textwidth}
         \centering
         \includegraphics[width=0.98\textwidth]{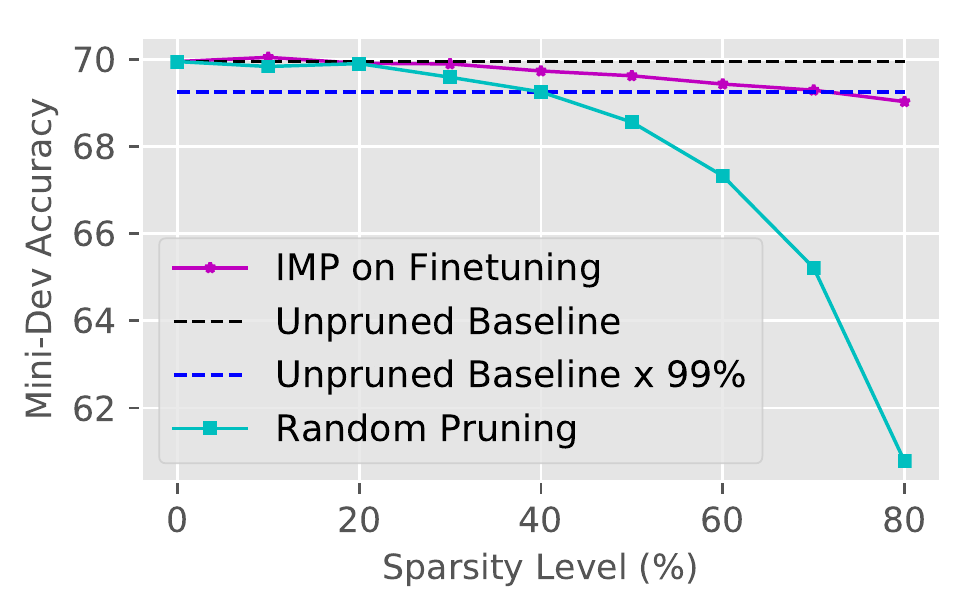}
         \caption{VQA}
     \end{subfigure}
     \begin{subfigure}[b]{0.32\textwidth}
         \centering
         \includegraphics[width=0.98\textwidth]{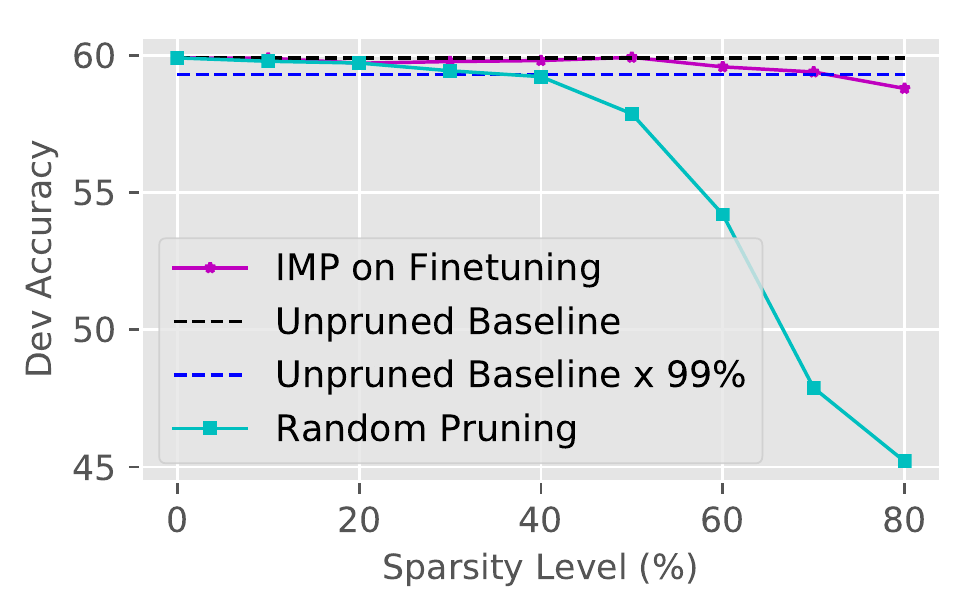}
         \caption{GQA}
     \end{subfigure}
     \begin{subfigure}[b]{0.32\textwidth}
         \centering
         \includegraphics[width=0.98\textwidth]{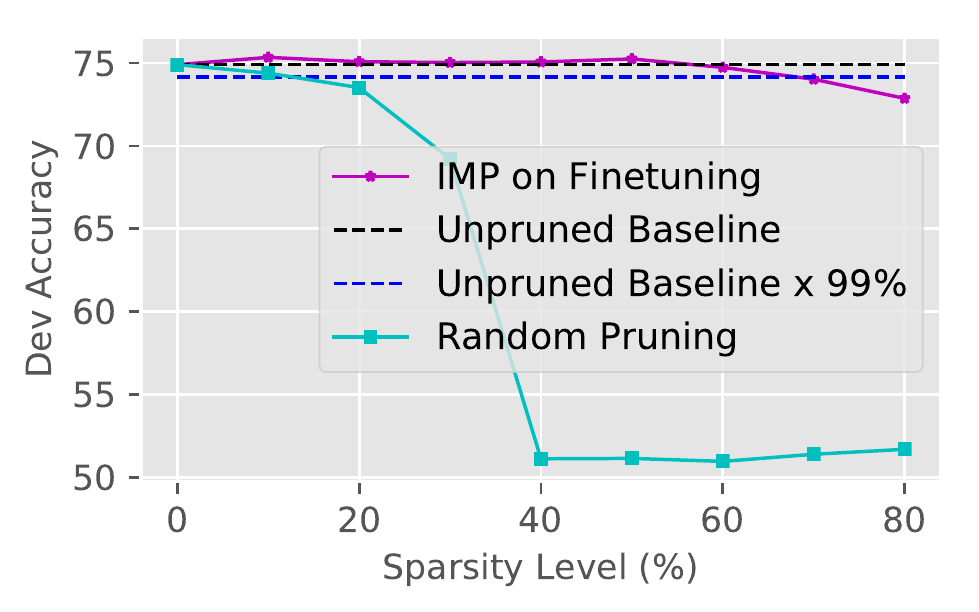}
         \caption{NLVR$^2$}
     \end{subfigure}
     \caption{The lottery ticket results of LXMERT on VQA, GQA, and NLVR$^2$.}
     \label{fig:lxmert_lottery}
\end{figure*}

\begin{figure*}[t]
     \centering
     \begin{subfigure}[b]{0.32\textwidth}
         \centering
         \includegraphics[width=0.98\textwidth]{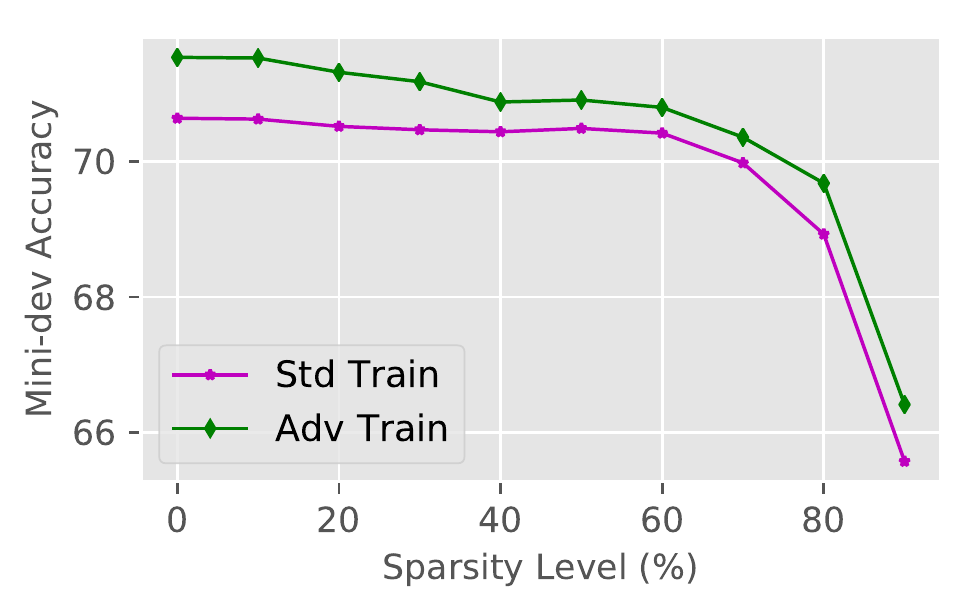}
         \caption{VQA}
     \end{subfigure}
     \begin{subfigure}[b]{0.32\textwidth}
         \centering
         \includegraphics[width=0.98\textwidth]{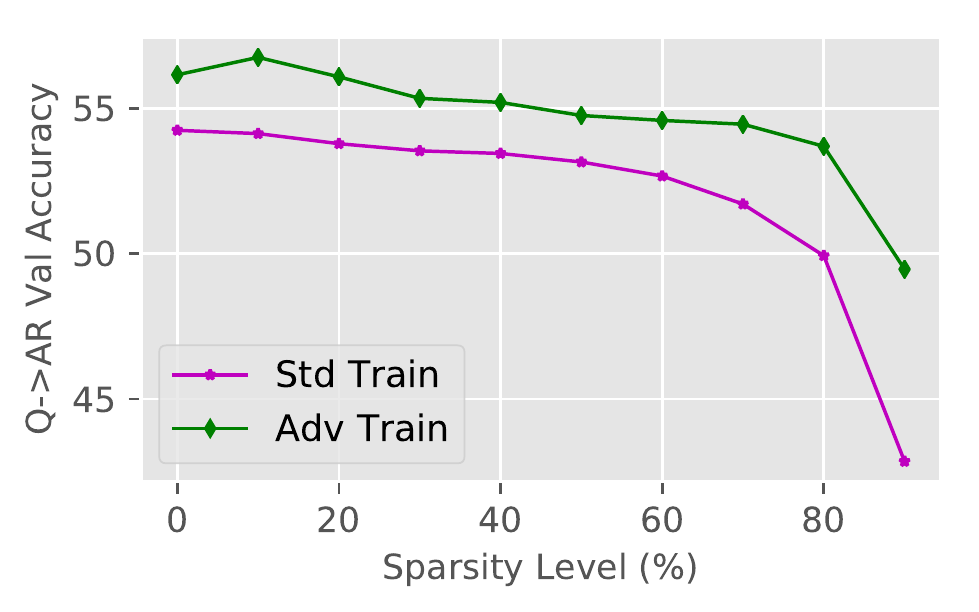}
         \caption{VCR}
    \end{subfigure}
    \begin{subfigure}[b]{0.32\textwidth}
         \centering
         \includegraphics[width=0.98\textwidth]{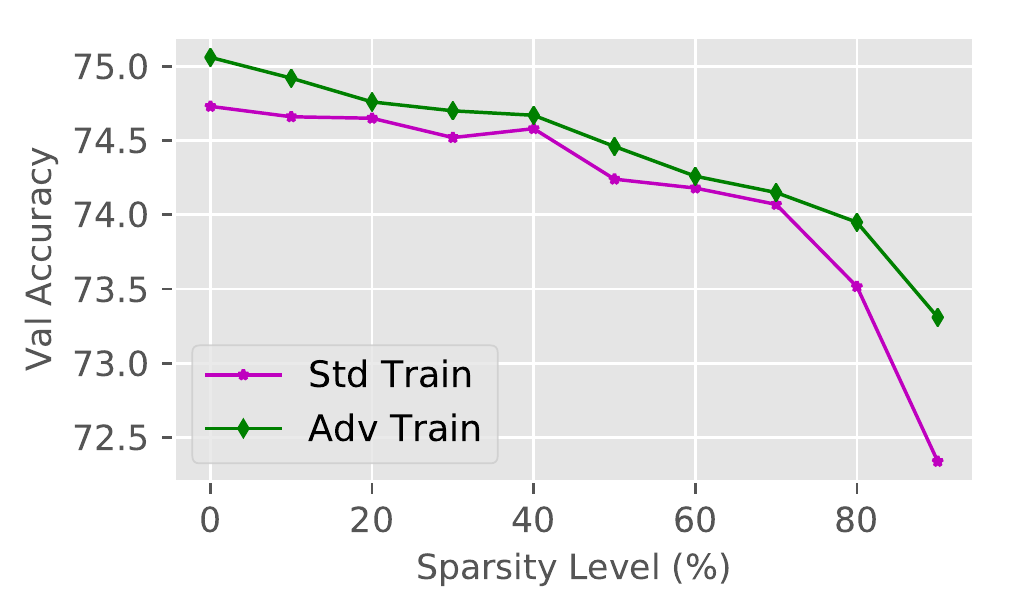}
         \caption{RefCOCO+}
     \end{subfigure}
     \caption{Performance of subnetworks that are found by adversarial training on the tasks of VQA, VCR and RefCOCO+.}
     \label{fig:adv_train_study}
\end{figure*}

\paragraph{\emph{Q5: Do winning tickets found on downstream tasks transfer?}}
One would also wonder whether such transfer learning behavior also exists among the downstream tasks themselves, \emph{i.e.}, whether the found subnetwork on a source task $\Scal$ transfers to a target task $\Tcal$. We perform a systematic study in Figure~\ref{fig:trans_study}, where within each plot, 8 ticket sources are considered. There are several key observations. ($i$) The subnetworks found by task-specific signals typically perform the best, especially on the high-sparsity regime. ($ii$) Surprisingly, all the individual subnetworks found by downstream tasks transfer well, indicating that models on all the tasks have learned some shared essential knowledge. ($iii$) The subnetwork from pre-training generally performs better than those from other tasks (\emph{e.g.}, 0.71\%-2.69\% better than other source tickets at 70\% sparsity on the VCR task), indicating its universal transferability. 
By taking a closer look at Figure~\ref{fig:trans_study}(a), excluding VQA itself, the best source ticket is from pre-training and GQA, as the task nature of VQA and GQA is similar. From Figure~\ref{fig:trans_study}(e) and (f), we can see the best source ticket for image-text retrieval is from pre-training. This is because the image-text matching task used in pre-training is similar to the downstream task itself. In Appendix, we also compare the similarity of sparsity patterns found on each downstream task.

Since subnetworks found on pre-training performs the best, we further compare their performance with the full model in more detail, and summarize results in Table~\ref{tab:universal_subnetwork_results}. The universal subnetwork at 60\%/70\% sparsity matches 98\%/96\%\footnote{This number changes to 99\%/97\% if  VCR is not counted in.} of the full accuracy over all the tasks considered, effectively serving as a task-agnositic compressed model. 

\begin{table}[!t]
\small
\begin{tabular}{c|ccc}
\multirow{2}{*}{Dataset} & VQA & GQA & NLVR$^2$  \\ 
& mini-dev$^\dagger$ & test-dev  & dev \\ \cline{2-4}
Sparsity & 70\% & 70\% & 70\% \\
\hline
LXMERT (paper) & 69.90 & 59.80 & 74.95 \\
LXMERT (reimp.) & 69.95$\pm$\tiny{0.03} & 59.91$\pm$\tiny{0.07} & 74.90$\pm$\tiny{0.26} \\
$\times 99\%$ & 69.25 & 59.31 & 74.15 \\ 
\hline
Lottery Tickets & 69.29$\pm$\tiny{0.10} & 59.40$\pm$\tiny{0.17} & 74.03$\pm$\tiny{0.71} \\
Random Pruning & 65.22$\pm$\tiny{0.05} & 47.88$\pm$\tiny{0.55} & 51.38$\pm$\tiny{0.45}  \\
\end{tabular}
\caption{
The LTH results of LXMERT on VQA, GQA, and NLVR$^2$. ($\dagger$) The same mini-dev set as used in LXMERT.    
}
\label{tab:lxmert_lottery_results}
\end{table}

\subsection{Additional Study}

\paragraph{\emph{Q6: Do different VLP models behave differently?}}
So far, we have focused on UNITER. Below, we experiment with LXMERT and ViLT to provide a more complete picture of VL lottery tickets. Results are summarized in Table~\ref{tab:lxmert_lottery_results},~\ref{tab:vilt_lottery_results}, and Figure~\ref{fig:lxmert_lottery}. For LXMERT, similar observations can be found. Since both UNITER and LXMERT use the same visual features from object detection, but only differ in the use of one-/two-stream architecture, we conclude that the LTH observations are not sensitive to this one-/two-stream design. On the other hand, ViLT can only achieve a low sparsity ratio (30\%) if we want to keep impaired performance. This is partially due to that ViLT directly takes image patches as input, all the modeling power needs to be absorbed in a single unified transformer, therefore less can be pruned, while for UNITER and LXMERT, the extracted image features are kept intact. 

\begin{table}[!t]
\small
\begin{tabular}{c|cc}
Dataset & VQA (mini-dev$^\dagger$)  & NLVR$^2$ (dev)  \\ 
\cline{2-3}
Sparsity & 30\%  & 30\% \\
\hline
ViLT (reimp.) & 70.88$\pm$\tiny{0.05} & 75.82$\pm$\tiny{0.20} \\
$\times 99\%$ & 70.17 & 75.06 \\ 
\hline
Lottery Tickets & 70.51$\pm$\tiny{0.11} & 75.22$\pm$\tiny{0.41} \\
Random Pruning & 65.16$\pm$\tiny{0.05} & 56.14$\pm$\tiny{0.40}  \\
\end{tabular}
\caption{
The lottery ticket results of ViLT on VQA and NLVR$^2$. ($\dagger$) The same mini-dev set as used in ViLT.    
}
\label{tab:vilt_lottery_results}
\end{table}

\paragraph{\emph{Q7: Can VLP models play lottery tickets adversarially?}} 
Lottery tickets are typically found via standard cross-entropy training. Here, we study whether adversarial training can be used to find winning tickets as well. Results are shown in Figure~\ref{fig:adv_train_study}. Interestingly, on the 3 tasks considered, the ticket performance via adversarial training at 80\% and 70\% sparsity matches (or almost matches) the performance via standard finetuning at 70\% and 60\% sparsity, respectively. This suggests that adversarial training has the effect of making the sparse winning tickets 10\% sparser in order to match the performance of a standard trained one. 

\section{Conclusion and Discussion}
In this paper, we have presented a comprehensive study of the lottery ticket hypothesis (LTH) for vision and language. Below, we discuss some limitations of the current study.
($i$) \emph{Efficiency}: We mainly focused on the scientific study of LTH. For future work, we plan to investigate the real speedup results on a hardware platform that is friendly to unstructured pruning, such as XNNPACK~\cite{elsen2020fast}. 
($ii$) \emph{Object Detection}: For UNITER/LXMERT, we studied the LTH for multimodal fusion, while keeping the object detection module untouched. In terms of end-to-end VLP, we focused on ViLT. For future work, we plan to study the LTH of object detection and other end-to-end VLP models. 

{\small
\bibliography{aaai22}
}

\clearpage
\appendix
\section{Details on Pruning and Adversarial Training}

\subsection{Unstructured Pruning}
We mainly use Iterative Magnitude-based Pruning (IMP) to find winning tickets. The pruning procedure is summarized in Algorithm~\ref{alg:IMP}.

\begin{algorithm}[h]
\caption{Iterative Magnitude Pruning for VL Tickets.}
\label{alg:IMP}
\begin{algorithmic}
    \STATE {\bfseries Input}\,\, Initial mask $\mv=1^{d_1}$; Pre-trained parameters $\thetav_0$ and task-specific parameters $\phiv_0$; rewinding step $i$ (could be 0), sparsity level $s$, total training step $t$.
    \STATE Train the pre-trained VL model $f(\xv; \mv \odot \thetav_0,\phiv_0)$ to step $i$: $f(\xv; \mv \odot \thetav_i,\phiv_i)$. 
    \REPEAT
    \STATE Train $f(\xv;  \mv \odot \thetav_i,\phiv_i)$ to step $t$: $f(\xv;  \mv \odot \thetav_t,\phiv_t)$. 
    \STATE Prune $10\%$ of non-zero weights of $\mv \odot \thetav_t$ based on the magnitudes and update $\mv$ accordingly.
    \UNTIL{the sparsity of $\mv$ reaches $s$}
    \STATE {\bfseries Return}\,\, $f(\xv; \mv \odot \thetav_i,\cdot)$
\end{algorithmic}
\end{algorithm}

\begin{figure*}[t!]
     \centering
     \begin{subfigure}[b]{0.4\textwidth}
         \centering
         \includegraphics[width=\textwidth]{figures/nlvr2_trans.pdf}
         \caption{VE}
     \end{subfigure}
     \begin{subfigure}[b]{0.4\textwidth}
         \centering
         \includegraphics[width=\textwidth]{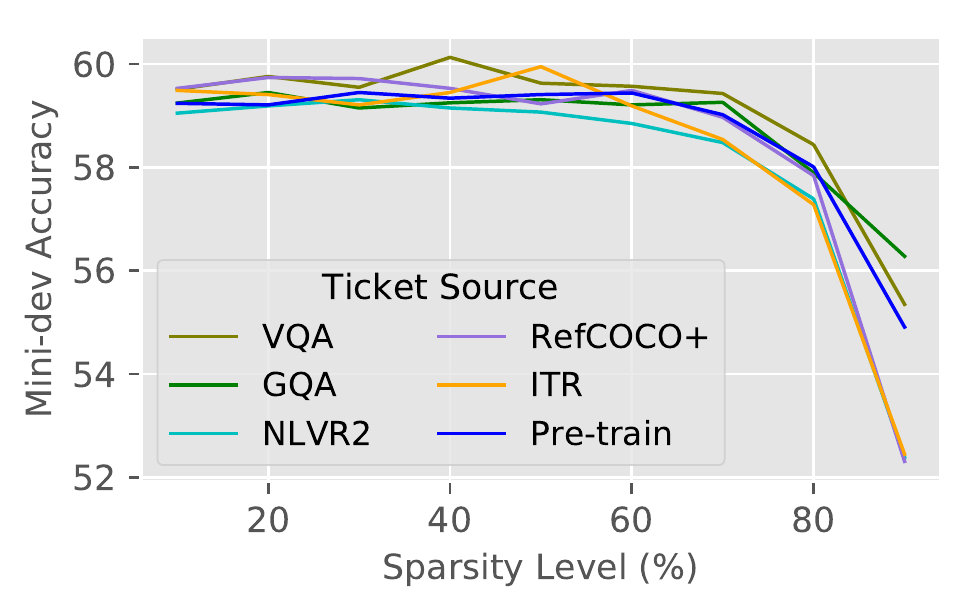}
         \caption{GQA}
     \end{subfigure}
     \caption{Transferring winning tickets across tasks. Winning ticket performance on target tasks: (a) VE, (b) GQA. Within each plot, each line represents a different source task for the winning ticket. Better zoomed in and viewed in color.}
     \label{fig:trans_study_supp}
\end{figure*}
\begin{figure*}[t!]
     \centering
     \begin{subfigure}[b]{0.4\textwidth}
         \centering
         \includegraphics[width=\textwidth]{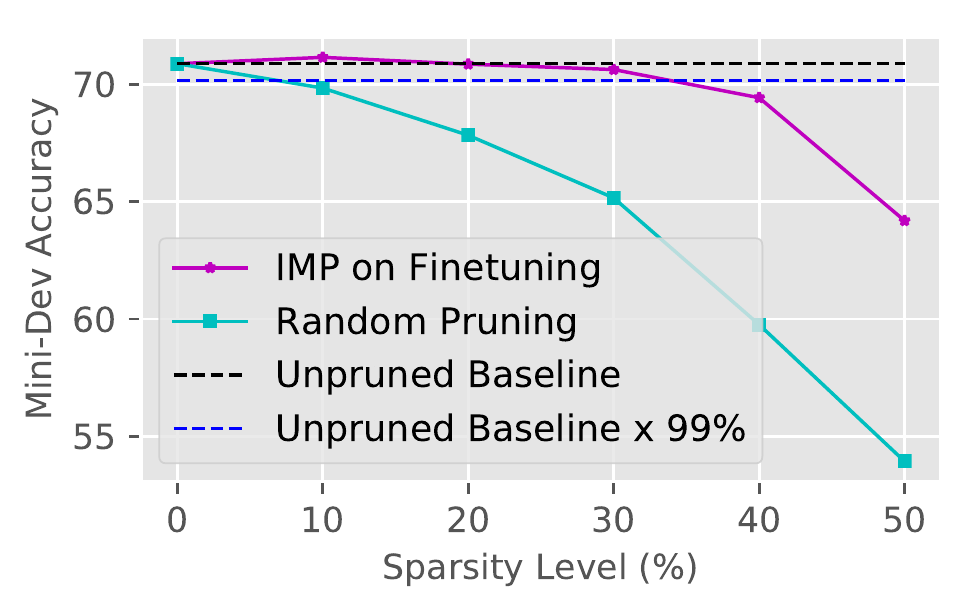}
         \caption{VQA}
     \end{subfigure}
     \begin{subfigure}[b]{0.4\textwidth}
         \centering
         \includegraphics[width=\textwidth]{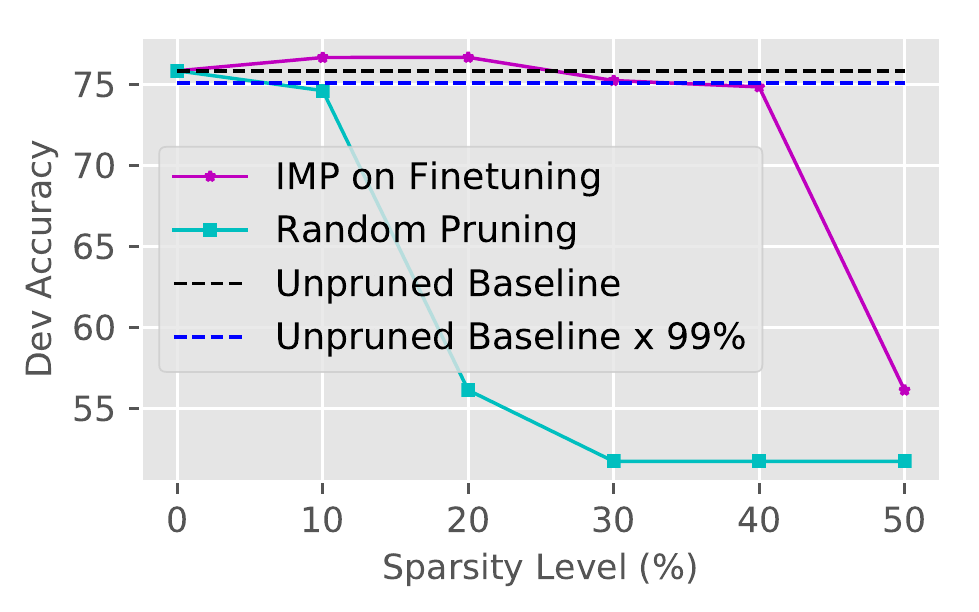}
         \caption{NLVR$^2$}
     \end{subfigure}
     \caption{The lottery ticket results of ViLT on VQA and NLVR$^2$.}
     \label{fig:vilt_lottery_curve}
\end{figure*}

\subsection{Lottery Tickets with Adversarial Training}
When adding adversarial perturbations into the feature space (\emph{e.g.}, image regional features and word embeddings), recent work~\cite{gan2020large} has shown that adversarial training (AT) can be used as an effective regularization to improve model performance. When combined with the found lottery tickets, the new objective becomes:
\begin{align} \label{eqn:outer_min}
    &\min_{\thetav,\phiv} \E_{(\xv,\yv)\sim \Dcal} \Big[ \Lcal_{std}(\xv,\yv;\mv \odot \thetav,\phiv) +  \\
    &\Rcal_{at} (\xv,\yv;\mv \odot \thetav,\phiv) + \alpha \cdot\Rcal_{kl} (\xv;\mv \odot \thetav,\phiv) \Big]\,,
\end{align}
where $\Lcal_{std}(\cdot)$ is the cross-entropy loss on clean data, $\Rcal_{at} (\cdot)$ is the label-preserving AT loss, and $\Rcal_{kl} (\cdot)$ is an adversarial regularization term. Specifically, 
\begin{align} \label{eqn:inner_max}
    \Rcal_{at} (\cdot) &= \max_{||\deltav||\leq \epsilon} L(f(\xv+\deltav;\mv \odot \thetav,\phiv),\yv)\,, \\
    \Rcal_{kl} (\cdot) &= \max_{||\deltav||\leq \epsilon} L_{kl}(f(\xv+\deltav;\mv \odot \thetav,\phiv),f(\xv;\mv \odot \thetav,\phiv))\,, \nonumber
\end{align}
where $L$ is the cross-entropy loss, $L_{kl}(p,q) = \mbox{KL}(p||q)+ \mbox{KL}(q||p)$, $p, q$ denote the two probability distributions, and $\mbox{KL}(\cdot)$ denotes the Kullback-Leibler divergence. Frobenius norm is used to constrain $\deltav$. For optimization, \citet{madry2017towards}~showed that the outer minimization in Eqn.(\ref{eqn:outer_min}) can be solved by SGD, while the inner maximization in Eqn.(\ref{eqn:inner_max}) can be solved reliably by projected gradient descent (PGD), which takes the following step (with step-size $\alpha$) in each iteration:
\begin{align} 
    \deltav_{t+1} = \Pi_{||\deltav||\leq \epsilon} (\deltav_{t}+\alpha g(\deltav_{t})/ ||g(\deltav_{t}) ||_F)\,,
\end{align}
where $g(\deltav_{t}) = \nabla_{\deltav}L(f(\xv+\deltav; \mv \odot \thetav, \phiv),\yv)$ is the gradient of the loss w.r.t. $\deltav$, and $\Pi_{||\deltav||\leq \epsilon}$ performs a projection onto the $\epsilon$-ball. Adversarial training is often used for dense neural netowrk training; here, we study the use of it for finding winning tickets.

\begin{figure*}[t!]
     \centering
     \includegraphics[width=0.45\linewidth]{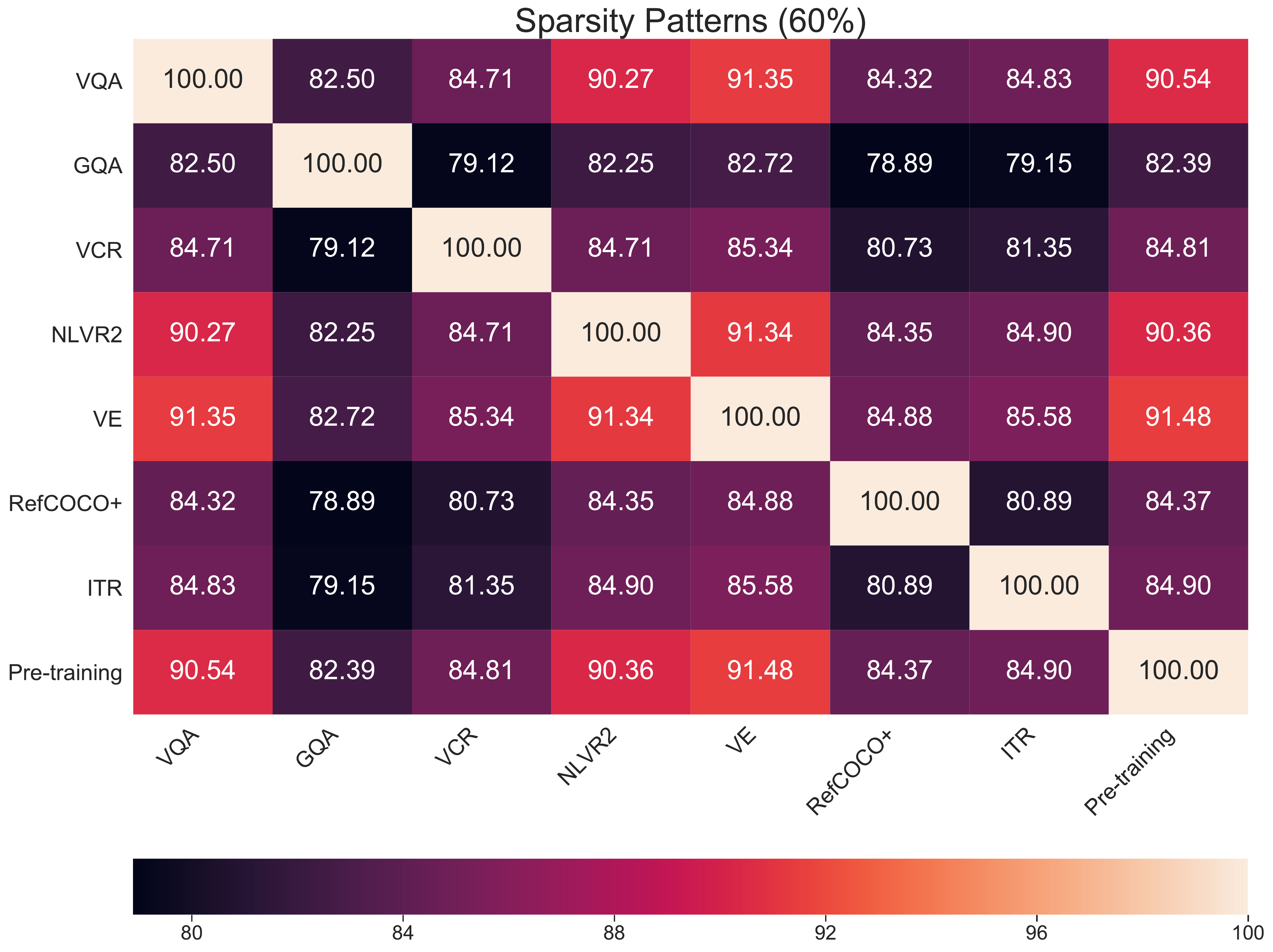}
     \includegraphics[width=0.45\linewidth]{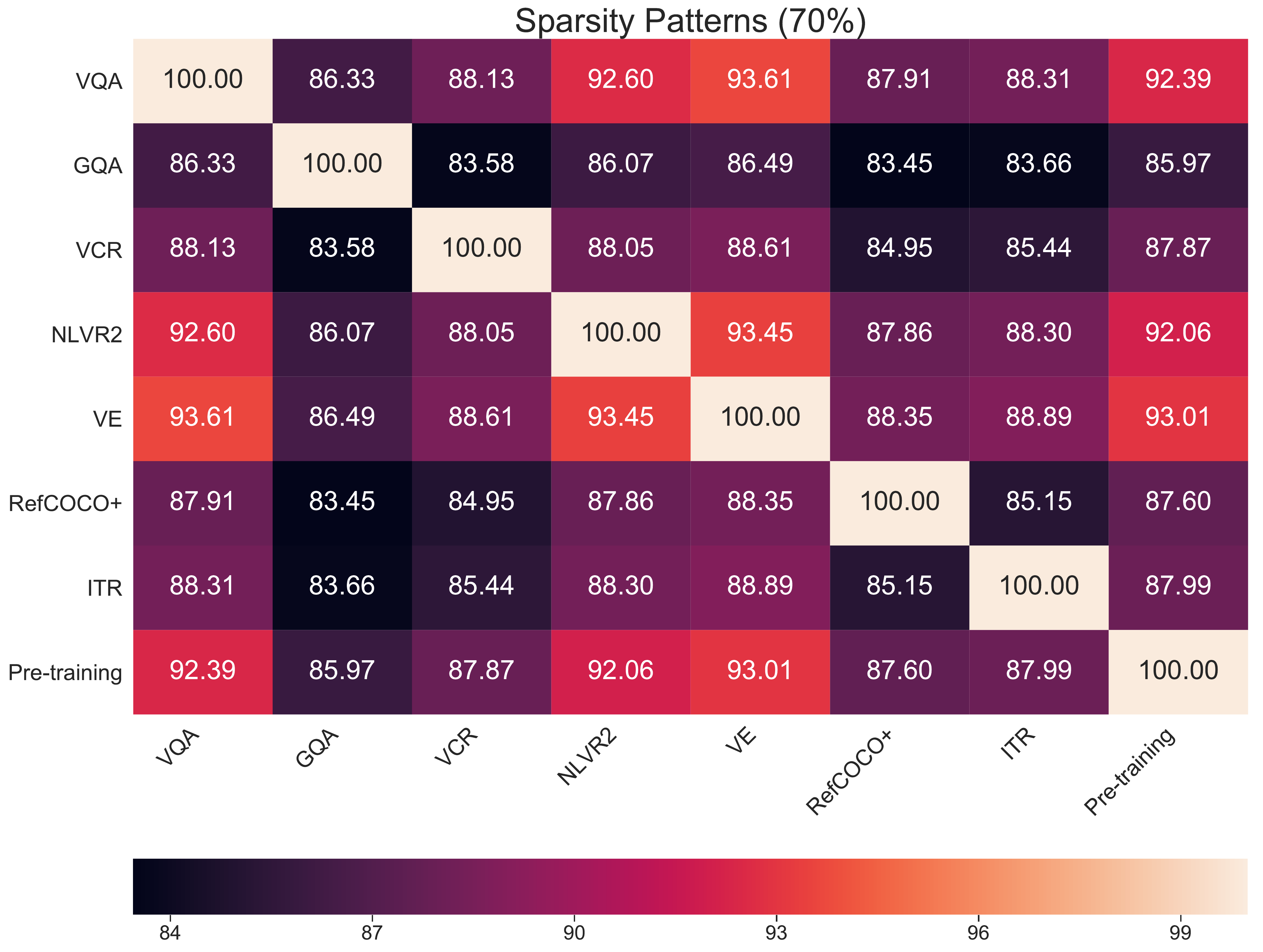}
     \caption{The overlap in sparsity patterns found on each downstream task and pre-training tasks with with sparsity 60\% and 70\%, respectively.}
     \label{fig:mask_similarity}
\end{figure*}

\subsection{Additional Results} \label{sec:add_results}

\paragraph{Additional Transfer Learning Study} 
In Figure~\ref{fig:trans_study_supp}, we show additional results on the rest two tasks that have not been covered in the main text: (a) VE, and (b) GQA. Consistently with the findings in the main text, all the ticket sources demonstrate very well transferability. For GQA, it is interesting to observe that the ticket source from VQA performs the best. This is intuitive to understanding, as the task nature of VQA and GQA is close, and VQA has a larger training dataset, therefore demonstrating better transferability, even better than the ticket soruce from GQA itself. 

\paragraph{Additional ViLT Lottery Ticket Curves}
We provide additional lottery ticket results of ViLT at all sparsity levels in Figure~\ref{fig:vilt_lottery_curve}. Since ViLT uses only a single unified transformer to directly take image patches and word tokens as model input, less can be pruned, and the highest sparsity we can achieve without impairing the performance is only around 30\%-40\%. 

\paragraph{Similarity Between Sparsity Patterns}
In Figure~\ref{fig:mask_similarity}, we compare the overlap in sparsity patterns found on each downstream task and the pre-training tasks. Each cell contains the relative overlap ratio (\emph{i.e.}, $\frac{\mv_i \bigcap \mv_j}{\mv_i \bigcup \mv_j}\%$) between masks (\emph{i.e.}, $\mv_i, \mv_j$) from task $\Tcal_i$ and $\Tcal_j$. We find that different masks have been learned under different tasks. The mask learned by pre-training shares the most similarity with masks learned from VQA, NLVR$^2$ and VE. For GQA, RefCOCO+, and ITR, the learned masks share the least similarity with others. 

\begin{table}[!t]
\small
\resizebox{0.47\textwidth}{!}
{
\begin{tabular}{c|cccccc}
Sparisty & VQA & GQA & VCR & NLVR$^2$ & VE & RefCOCO+   \\ \cline{2-7}
\hline
60\% (Std.)  & 70.41 & 59.44 & 50.37 & 75.52 & 77.79 & 74.41  \\
60\% (Adv.)  & 70.80 & 59.85 & 51.07 & 76.70 & 77.99 & 74.74 \\
\hline
70\% (Std.) & 69.45 & 59.02 & 47.52 & 74.29 & 77.34 & 73.45 \\
70\% (Adv.) & 69.79 & 59.37 & 48.50 & 75.29 & 77.51 & 74.08 \\
\end{tabular}
}
\caption{
Performance of adversarial training on the universal subnetworks at 60\% and 70\% sparsities. Std.: standard cross-entropy training; Adv.: adversarial training.
}
\label{tab:adv_results}
\end{table}

\paragraph{Enhancing Winning Tickets with AT} 
Since the universal subnetworks found on pre-training
at 60\% and 70\% sparsities are the most interesting, we finetune them via adversarial training, and summarize results in
Table~\ref{tab:adv_results}. Clearly, performance on all the tasks is improved,
demonstrating adversarial training is also useful to enhance
sparse neural network training, at least in our VL context.

\end{document}